\documentclass{article}

\usepackage{arxiv}

\usepackage[utf8]{inputenc} % allow utf-8 input
\usepackage[T1]{fontenc}    % use 8-bit T1 fonts
\usepackage{hyperref}       % hyperlinks
\usepackage{url}            % simple URL typesetting
\usepackage{booktabs}       % professional-quality tables
\usepackage{amsmath}       % blackboard math symbols
\usepackage{nicefrac}       % compact symbols for 1/2, etc.
\usepackage{microtype}      % microtypography
\usepackage{lipsum}
\usepackage{multirow,bm}
\usepackage{graphicx}
\usepackage{geometry}
\usepackage{array}
\usepackage{algorithm,algorithmic}
\usepackage{xcolor}
\graphicspath{ {./images/} }

\usepackage{titlesec}
\titleformat{\paragraph}
{\normalfont\normalsize\bfseries}{\theparagraph}{1em}{}
\titlespacing*{\paragraph}
{0pt}{3.25ex plus 1ex minus .2ex}{1.5ex plus .2ex}

\title{Uncertainty Quantification Techniques for Space Weather Modeling: Thermospheric Density Application}

\author{
 Richard J. Licata \\
  Dept. of Mechanical and Aerospace Engineering\\
  West Virginia University\\
  Morgantown, WV 26505 \\
  \texttt{rjlicata@mix.wvu.edu} \\
   \And
 Piyush M. Mehta \\
  Dept. of Mechanical and Aerospace Engineering\\
  West Virginia University\\
  Morgantown, WV 26505 \\
}

\begin{document}
\maketitle
\begin{abstract}
Machine learning (ML) has been applied to space weather problems with increasing frequency in recent years, driven by an influx of in-situ measurements and a desire to improve modeling and forecasting capabilities throughout the field. Space weather originates from solar perturbations and is comprised of the resulting complex variations they cause within the numerous systems between the Sun and Earth. These systems are often tightly coupled and not well understood. This creates a need for skillful models with knowledge about the confidence of their predictions. One example of such a dynamical system highly impacted by space weather is the thermosphere, the neutral region of Earth's upper atmosphere. Our inability to forecast it has severe repercussions in the context of satellite drag and computation of probability of collision between two space objects in low Earth orbit (LEO) for decision making in space operations. Even with (assumed) perfect forecast of model drivers, our incomplete knowledge of the system results in often inaccurate thermospheric neutral mass density predictions. Continuing efforts are being made to improve model accuracy, but density models rarely provide estimates of confidence in predictions. In this work, we propose two techniques to develop nonlinear ML regression models to predict thermospheric density while providing robust and reliable uncertainty estimates: Monte Carlo (MC) dropout and direct prediction of the probability distribution, both using the negative logarithm of predictive density (NLPD) loss function. We show the performance capabilities for models trained on both local and global datasets. We show that the NLPD loss provides similar results for both techniques but the direct probability distribution prediction method has a much lower computational cost. For the global model regressed on the Space Environment Technologies High Accuracy Satellite Drag Model (HASDM) density database, we achieve errors of approximately 11\% on independent test data with well-calibrated uncertainty estimates. Using an in-situ CHAllenging Minisatellite Payload (CHAMP) density dataset, models developed using both techniques provide test error on the order of 13\%. The CHAMP models -- on validation and test data -- are within 2\% of perfect calibration for the twenty prediction intervals tested. We show that this model can also be used to obtain global density predictions with uncertainties at a given epoch.
\end{abstract}

\section{Introduction}

Low Earth orbit (LEO) will see the addition of tens of thousands of satellites in the coming years as private companies are developing mega constellations for the new space economy \cite{constellations}. This congestion of certain orbital regimes increases the likelihood of a future collision between two objects. Satellite collisions can create debris clouds consisting of thousands of objects large enough to pose significant threats to other space assets. The 2009 Iridium-Cosmos collision resulted in approximately 2,300 observable debris objects, 65\% of which remained in orbit seven years later \cite{iridium_cosmos}. Debris objects created by collisions or weapons tests can catapult into highly elliptical orbits which pose a danger to satellites in multiple orbital regimes \cite{constellations2021}.

In an effort to prevent these events from occurring, objects are continuously tracked, and their trajectories are predicted. However, uncertainties play a large role in the prediction of future satellite positions. In LEO, atmospheric drag is the largest single source of uncertainty mainly due to an incomplete understanding of the thermosphere. Variations in the thermosphere are connected to temperature changes, as the atmosphere expands and contracts. Solar extreme ultraviolet (EUV) and far ultraviolet (FUV) irradiance are the primary heating sources \cite{energetics_therm}. This absorption of solar irradiance provides the baseline thermospheric mass density \cite{therm_var}. The effects of solar emissions are well-represented by various solar indices and proxies \cite{dev_inds}. 

Solar irradiance is generally a long-term variation while the solar wind drives more rapid changes in the thermosphere. Mass and energy from the sun -- manifested as the solar wind -- travel through space and interact with the near-Earth geospace environment. Certain events (e.g. coronal mass ejections) send massive amounts of energy and mass that result in significant increases in thermospheric density. Energy, and therefore density, enhancements first appear in the auroral zone (high latitudes) and propagate towards the equator in the form of traveling atmospheric disturbances \cite{TAD1}. Geomagnetic storms are a particularly difficult phenomena to model and our current density models carry high uncertainty during these periods \cite{storms_CME,therm_storm}.

Satellite accelerometers have provided a unique insight into the thermosphere with high fidelity in-situ measurements, particularly during storms \cite{TAD2}. Accelerations caused by non-drag sources (e.g. gravity and solar radiation pressure) are modeled out allowing the isolation of drag acceleration that is then used to estimate mass density \cite{CHden1,CHden2,Sutton,Doorn,EOF3}. Drag acceleration is given as
%%%%%
\begin{equation} \label{eq1}
\begin{split}
\vec{a}_{\textrm{drag}} = -\frac{1}{2} \rho \frac{c_D A}{m} v^2_{\textrm{rel}}\frac{v_{\textrm{rel}}}{\left|v_{\textrm{rel}}\right|}
\end{split}
\end{equation}
%%%%%
where $\vec{a}_{\textrm{drag}}$ is the drag acceleration, $\rho$ is local mass density, $c_D$ is the satellite drag coefficient, $A$ is the cross-sectional area, $m$ is the satellite mass, and $v_{\textrm{rel}}$ is the relative velocity of the satellite with respect to the rotating atmosphere. With an estimate for drag acceleration, the density can be estimated, assuming adequate knowledge of the drag coefficient and cross-sectional area given the satellite orientation. Density estimates obtained through this method are considered ground truth and often used for model validation. 

Accelerometer and orbit-derived densities have been used frequently in developing empirical models \cite{NRLMSISE00,JB2008,DTM}. Furthermore, they have been used in data assimilation schemes to make corrections to background models, either through observed orbital drag data \cite{suttonDC} or two-line element data \cite{doornDC}. The most prominent integration of real-time data for neutral density modeling is the High Accuracy Satellite Drag Model's (HASDM) Dynamic Calibration of the Atmosphere (DCA). This uses observed satellite data to make corrections to a background empirical density model \cite{HASDM}.

Even with these improvements, density models have high errors, and we generally use them without any knowledge of their confidence given the conditions. Until recently, no thermospheric density models -- whether physics-based or empirical-- provided estimates of uncertainty. Bruinsma et al. (2021) developed an uncertainty-based version of DTM2020 using polynomials to describe the 1-sigma uncertainties as a function of the inputs \cite{DTMUQ}. Licata et al. (2021) used MC dropout to obtain uncertainty estimates for a global density modeling application with good calibration, providing baseline performance \cite{HASDM-ML}. In this work, we leverage ML to generate predictive density models for the thermosphere that also provide robust and reliable uncertainty estimates. This is done for both a global and local datasets using two methods: Monte Carlo (MC) dropout and a direct prediction of the probability distribution (referred to primarily as direct probability). 

We first outline the data and methods used for model development and analysis. Then, we use artificial data to demonstrate the techniques. We move to the results for modeling with the global density dataset using the two uncertainty techniques and perform a similar analysis for the models developed on local measurements. We also look at the global prediction capabilities of the model developed with in-situ data, and we compare the evaluation times of both uncertainty methods.

\section{Datasets}

\subsection{SET HASDM Density Database}

The High Accuracy Satellite Drag Model (HASDM) is the operational thermospheric density framework used by the United States Space Force (USSF) \cite{HASDM}. By improving the density correction techniques presented by Marcos et al. (1998) \cite{Marcos} and Nazerenko et al. (1998) \cite{Naz}, HASDM modifies 13 global temperature coefficients to make real-time corrections to the Jacchia-Bowman 2008 Empirical Thermospheric Density Model (JB2008) \cite{JB2008}. Its Dynamic Calibration of the Atmosphere (DCA) algorithm ingests data from calibration satellites that are distributed in altitude between 190 and 900 km -- most are between 300 and 600 km \cite{HASDMrev}. As the algorithm provides corrections to JB2008, HASDM provides global measurements on a 24$\times$19$\times$27 grid. For additional information on HASDM, the reader is referred to Storz et al. (2005).

While HASDM is highly desired due to its real-time data assimilation, it is a proprietary model that is inaccessible to researchers and operators. Space Environment Technologies (SET) is the contractor responsible for validating HASDM outputs on a weekly basis, and they recently released HASDM validation archives from 2000-2020 covering close to two full solar cycles providing good statistical coverage. %add sentence about solar cycle
This data release constitutes the SET HASDM density database \cite{HASDM_SET_Data}. With a three-hour cadence, the database contains 58,440 global HASDM outputs. Each output has a resolution of 15$^\circ$ longitude, 10$^\circ$ latitude, and 25 km altitude ranging from 175 to 825 km. For further details on the SET HASDM density database and the validation process, the reader is referred to Tobiska et al. (2021).

\subsection{Satellite Accelerometer Density Estimates} \label{sec:CHAMP}

\begin{figure}[htb!]
	\centering
	\small
	\includegraphics[width=\textwidth]{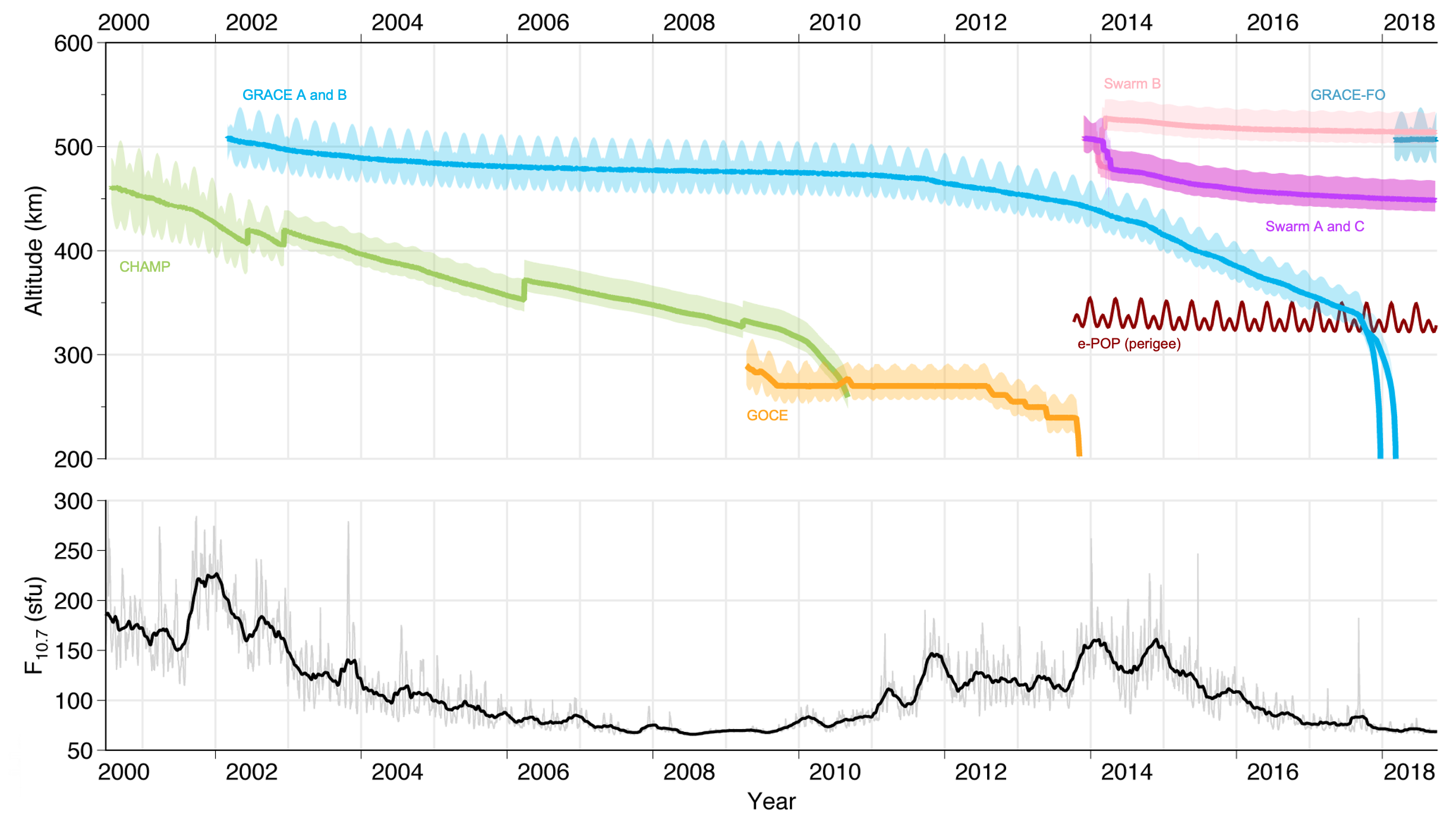}
	\caption{Timeline of satellites with onboard accelerometers from 2000 -- 2018 with bold lines representing the satellites' mean altitude. Note: e-POP does not contain an accelerometer. The bottom panel shows the corresponding \textit{F\textsubscript{10}} values indicating solar activity. The authors gratefully acknowledge Dr. Eelco Doornbos for providing this figure.}
	\label{f:sats}
\end{figure}

CHAllenging Minisatellite Payload (CHAMP) was launched in mid-2000 to study Earth's gravity and magnetic fields \cite{CHAMP}. Its orbit is nearly polar with an inclination of 87.3$^\circ$ providing adequate global coverage, and it began at 460 km in altitude. CHAMP was in orbit until 2010 when it stopped providing measurements at an altitude of approximately 300 km. The long mission lifetime covered nearly a solar cycle, providing measurements in solar maximum across many strong geomagnetic storms and through the following solar minimum. Mehta et al. (2017) used higher fidelity satellite geometry and improved gas-surface interaction models to scale the CHAMP density estimates of Sutton (2008) \cite{CHAGRA,DSMC,cd1,cd2}. The density dataset starts on January 1, 2002 and ends on February 22, 2010 with a 10-second cadence. The CHAMP dataset is prime for demonstration as spatiotemporally limited in-situ datasets are common in the space weather field. This type of model can be built upon with the addition of density estimates from other satellites, displayed in Figure \ref{f:sats}.

The addition of all satellites shown is Figure \ref{f:sats} would significantly expand the altitude coverage of the in-situ density dataset. The CHAMP dataset used in this work has a 160 km altitude range and does not span a full solar cycle. Integrating the density datasets of Gravity Recovery and Climate Experiment (GRACE) \cite{GRACE}, Gravity Field and Steady-State Ocean Circulation Explorer (GOCE) \cite{GOCE}, and Swarm \cite{Swarm} would provide a thorough altitude coverage of approximately 220 -- 550 km and span from 2001 to present day. The Enhanced Polar Outflow Probe (e-POP) \cite{EPOP} is a payload on Cascade, Smallsat and Ionospheric Polar Explorer (CASSIOPE) and its density estimates can be obtained through the processing of its Global Navigation Satellite System (GNSS) receivers \cite{CASSIOPE1,CASSIOPE2}. However, there is still much active research related to the proper combination of different satellite density datasets \cite{intercal,MarchA,calibration1}. Therefore, we proceed with the standalone CHAMP dataset for demonstration.

\subsection{Density Model Drivers}

JB2008 uses four solar indices/proxies as drivers for solar activity. \textit{F\textsubscript{10}} -- more completely referred to as \textit{F\textsubscript{10.7}} -- represents 10.7 cm solar radio flux and is a reliable proxy for solar EUV heating. \textit{S\textsubscript{10}} is an index for the integrated 26-34 nm solar EUV emission. The \textit{M\textsubscript{10}} proxy is a surrogate for FUV photospheric 160-nm Schumann-Runge Continuum emissions. \textit{Y\textsubscript{10}} is a hybrid index that measures solar coronal X-ray emissions during solar maximum and Lyman-$\alpha$ emissions during solar minimum. The \textit{S\textsubscript{10}}, \textit{M\textsubscript{10}}, and \textit{Y\textsubscript{10}} indices and proxies are not related to the 10.7 cm wavelength, but they are converted to \textit{F\textsubscript{10}} units -- solar flux units (sfu) -- through linear regression. JB2008 also uses the 81-day centered averages for all four solar drivers. This is indicated by the "\textit{81c}" subscript. Additional information on these solar drivers is provided by Tobiska et al. (2008).

To model geomagnetic activity, JB2008 uses a combination of \textit{a\textsubscript{p}} and \textit{Dst}. The \textit{a\textsubscript{p}} index represents global geomagnetic activity with a three-hour cadence. While it is widely used in density models, it is limited by the low-latitude range of the measurements and its discrete range of 28 values. \textit{Dst} is an index driven by the ring current strength in the inner magnetosphere \cite{ring_current}. When \textit{Dst\textsubscript{min}} is below -75 \textit{nT}, JB2008 shifts to using \textit{Dst} as it improves storm-time performance \cite{JB2008}.  The EXTEMPLAR (EXospheric TEMperatures on a PoLyhedrAl gRid) model uses Poynting flux totals in the northern and southern hemispheres -- \textit{S\textsubscript{N}} and \textit{S\textsubscript{S}}, respectively \cite{EXTEMPLAR}. Poynting flux represents electrodynamic energy flowing into the upper atmosphere. The \textit{a\textsubscript{p}} and \textit{Dst} indices have 3-hour and 1-hour cadences, respectively. Therefore, their use in a high-cadence model would not be advised. The geomagnetic index used to replace \textit{a\textsubscript{p}} and \textit{Dst} in the CHAMP model is \textit{SYM-H}, the longitudinally symmetric component of the magnetic field disturbances \cite{sym1,sym2}. \textit{SYM-H} is available with a 1-minute cadence.

The input sets for the HASDM and CHAMP models are shown in Table \ref{t:inputs}. The transformed time inputs \textit{t\textsubscript{1}} - \textit{t\textsubscript{4}} are defined in Equation \ref{eqT}, and the transformed local solar time (LST) inputs are defined in Equation \ref{eqLST}. These transformations are performed to make the time and location inputs continuous. 

\begin{equation} \label{eqT}
\begin{split}
t_1=sin\left(\frac{2\pi doy}{365.25}\right),
\;\;\;\;
t_2=cos\left(\frac{2\pi doy}{365.25}\right),
\;\;\;\;
t_3=sin\left(\frac{2\pi UT}{24}\right),
\;\;\;\;
t_4=cos\left(\frac{2\pi UT}{24}\right).
\end{split}
\end{equation}

\begin{equation} \label{eqLST}
\begin{split}
LST_1=sin\left(\frac{2\pi LST}{24}\right)
\;\;\;\;
LST_2=cos\left(\frac{2\pi LST}{24}\right)
\end{split}
\end{equation}

\begin{table}[htb!]
	\fontsize{10}{10}\selectfont
    \caption{List of inputs for both models.}
   \label{t:inputs}
        \centering 
   \begin{tabular}{c | c | c | c | c | c } % Column formatting, 
      \hline 
          \multicolumn{3}{c |}{\textbf{HASDM}} & \multicolumn{3}{c }{\textbf{CHAMP}} \\ \hline
          \textbf{Solar} & \textbf{Geomagnetic} & \textbf{Temporal} & \textbf{Solar} & \textbf{Geomagnetic} & \textbf{Spatial/Temporal} \\ \hline 
          $F_{10}$,  $S_{10}$, & $a_{pA}$,  $a_p$,  $a_{p3}$, & $t_1$,  $t_2$, & $F_{10}$,  $S_{10}$, & \textit{SYM-H}, & $LST_1$, $LST_2$, \\
          $M_{10}$,  $Y_{10}$, & $a_{p6}$,  $a_{p9}$,  $a_{p12-33}$, & $t_3$,  $t_4$ & $M_{10}$,  $Y_{10}$, & $S_N$, $S_S$ & $LAT$, $ALT$, \\
          $F_{81c}$,  $S_{81c}$, & $a_{p36-57}$,  \textit{Dst}$_A$, \textit{Dst}, & & $F_{81c}$,  $S_{81c}$, & & $t_1$, $t_2$, \\
          $M_{81c}$,  $Y_{81c}$ & \textit{Dst}$_3$, \textit{Dst}$_6$,  \textit{Dst}$_9$, & & $M_{81c}$,  $Y_{81c}$ & & $t_3$, $t_4$\\
          & \textit{Dst}$_{12}$, \textit{Dst}$_{15}$, \textit{Dst}$_{18}$, \textit{Dst}$_{21}$ & & &  & \\ \hline
   \end{tabular}
\end{table}

In Table \ref{t:inputs}, \textit{LST}, \textit{LAT}, and \textit{ALT} are the local time, latitude, and altitude of the satellite, respectively. The "A" subscript for the geomagnetic indices refers to the daily average. The numerical subscripts for these indices refer to the value that many hours prior to prediction epoch. The combination of numbers refers to the average of that index over that many hours prior to epoch. The HASDM input set originates from Licata et al. (2021) \cite{HASDM_ML}.

\section{Methodology}
\subsection{Machine Learning}
\subsubsection{Principal Component Analysis}

Principal Component Analysis (PCA), also referred to as Empirical Orthogonal Function (EOF) analysis, has been widely used in thermospheric mass density applications. It has been applied to satellite accelerometer datasets (as described in Section \ref{sec:CHAMP}) to identify dominant modes of variability in the thermosphere \cite{EOF1, EOF2, EOF3}. PCA is also used in the field for dimensionality reduction as part of a reduced-order model (ROM) \cite{MehtaROMCal,MehtaROM,TLE_ROM}. The HASDM dataset has 12,312 model outputs each epoch which makes uncertainty quantification (UQ) infeasible. Therefore, we apply PCA to the dataset for ROM development with the goal of UQ. PCA is an eigendecomposition technique that maximizes variance, determining uncorrelated linear combinations of data \cite{Pearson_PCA,PCA2}. We first take the common logarithm (\textit{log\textsubscript{10}}) of the density values to reduce the data's variance then remove the spatial mean. PCA decomposes the data, separating the spatial and temporal variations such that
%%%%%
\begin{equation} \label{eq1}
\begin{split}
\mathbf{x}\left(\mathbf{s},t\right)=\mathbf{\bar{x}}\left(\mathbf{s}\right)+\mathbf{\widetilde{x}}\left(\mathbf{s},t\right)
\;\;\;\textrm{and}\;\;\;
\mathbf{\widetilde{x}}\left(\mathbf{s},t\right)=\sum^r_{i=1}\alpha_i\left(t\right)U_i\left(s\right)
\end{split}
\end{equation}
%%%%%
In Equation \ref{eq1}, $\mathbf{x}\left(\mathbf{s},t\right)$ is the log density from HASDM, $\mathbf{\bar{x}}$ is the spatial mean, and $\mathbf{\widetilde{x}}$ is the variation about the mean. $\alpha_i(t)$ are temporal PCA coefficients and $U_i$ are orthogonal modes -- also called basis functions. The orthogonal modes are derived through
%%%%%
\begin{equation} \label{eq2}
\begin{split}
\mathbf{X} = U \Sigma V^T
\;\;\;\textrm{where}\;\;\;
\mathbf{X}=\left[
  \begin{array}{ccccc}
    \vrule & \vrule & \vrule &        & \vrule \\
    \mathbf{\widetilde{x}_1}    & \mathbf{\widetilde{x}_2} & \mathbf{\widetilde{x}_3}  & \ldots & \mathbf{\widetilde{x}_m}    \\
    \vrule & \vrule & \vrule &        & \vrule 
  \end{array}
\right]
\end{split}
\end{equation}
%%%%%
\textit{U} consists of orthogonal vectors representing the modes of variability. The $\Sigma$ matrix contains the squares of the eigenvalues -- corresponding to the columns in \textit{U} -- along the diagonal. The data is encoded by performing matrix multiplication with \textit{U}.

\subsubsection{Neural Network Modeling}

%justify our use of NNs
In this work, we leverage neural networks (NNs) for nonlinear regression modeling due to their applicability as universal function approximators and flexibility in development. A neural network is a collection of computational cells (or neurons) connected in some form through multiplicative connections (or weights). Neural networks were first conceived by McCulloch and Pitts (1943) when they described a computational representation of brain neurons and synapses with calculus and statistical theory \cite{NN_orig}. In the late 1950's, the first artificial neural network (ANN) was developed and is known as the perceptron \cite{perceptron}. Backpropagation is the process in which the network parameters are updated based on observations and was fundamental to the development of modern neural networks \cite{BP1,BP2}. ANNs have been used to directly predict thermospheric mass density using space weather indices and proxies as model drivers in order to study long-term trends \cite{ANNdensityLT1,ANNdensityLT2}. These types of models can also be used as an exercise in understanding the effect of the drivers on non-machine learning (ML) models \cite{MLthermEX}. Chen et al. (2014) \cite{ML_storm_dens} developed ANNs with different combinations of geomagnetic indices to fit to CHAMP and GRACE density estimates during storms, and Choury et al. (2013) \cite{TexANN} developed an ANN to predict exospheric temperature for use in the Drag Temperature Model (DTM).

\paragraph{Loss Functions} \label{sec:loss}

Loss functions are used to inform the NN of the objective during the training phase, or weight adjustment period. Loss functions can be minimized or maximized depending on the modeling objective. A common loss function in regression modeling is mean square error (MSE) given as,
%%%%%
\begin{equation} \label{eqMSE}
\begin{split}
MSE(y,\hat{y}) = \frac{1}{n}\sum^n_{i=1}\left(y_i - \hat{y}_i   \right)^2
\end{split}
\end{equation}
%%%%%
where \textit{y} is the ground truth, $\hat{y}$ is the model prediction, and \textit{n} is the batch size. The batch size is the number of samples the model will pass through before updating the weights, averaging the loss over the batch. Losses are computed for every output and backpropagation is how the model determines how much to change each weight. MSE is not used in this work (explained in Section \ref{sec:HPO}). We instead use the negative logarithm of predictive density (NLPD), 
%%%%%
\begin{equation} \label{eqNLPD}
\begin{split}
NLPD(y,\mu,\sigma) = \frac{\left(y-\mu\right)^2}{2\sigma^2} + \frac{ln(\sigma^2)}{2} + \frac{ln(2\pi)}{2}
\end{split}
\end{equation}
%%%%%
where \textit{y} is the observed value, $\mu$ is the predicted mean, and $\sigma$ is the standard deviation of the output corresponding to each unique input. NLPD is derived from $-ln(f(x))$ where $ln$ is the natural logarithm and $f(x)$ is the probability density function of the normal distribution.

\paragraph{Hyperparameter Optimization}\label{sec:HPO}

Tools like Keras Tuner have drastically reduced model development time \cite{KT}. You can provide ranges of hyperparameters and Keras Tuner will explore the search space. We use the Bayesian optimization scheme, allowing the tuner to perform a random search for the first 25 trials, or architectures, and using a Gaussian process model to choose the architectures for the final 75 trails to exploit the high performing areas of the space. The objective of the tuner is to minimize validation loss. The model optimizer and number of layers are first chosen by the tuner. For each layer, the model can have a unique number of neurons, activation function, and dropout rate. For each model developed (two datasets and two UQ techniques), the architecture is selected using Keras Tuner.

\begin{figure}[htb!]
	\centering
	\small
	\includegraphics[width=0.95\textwidth]{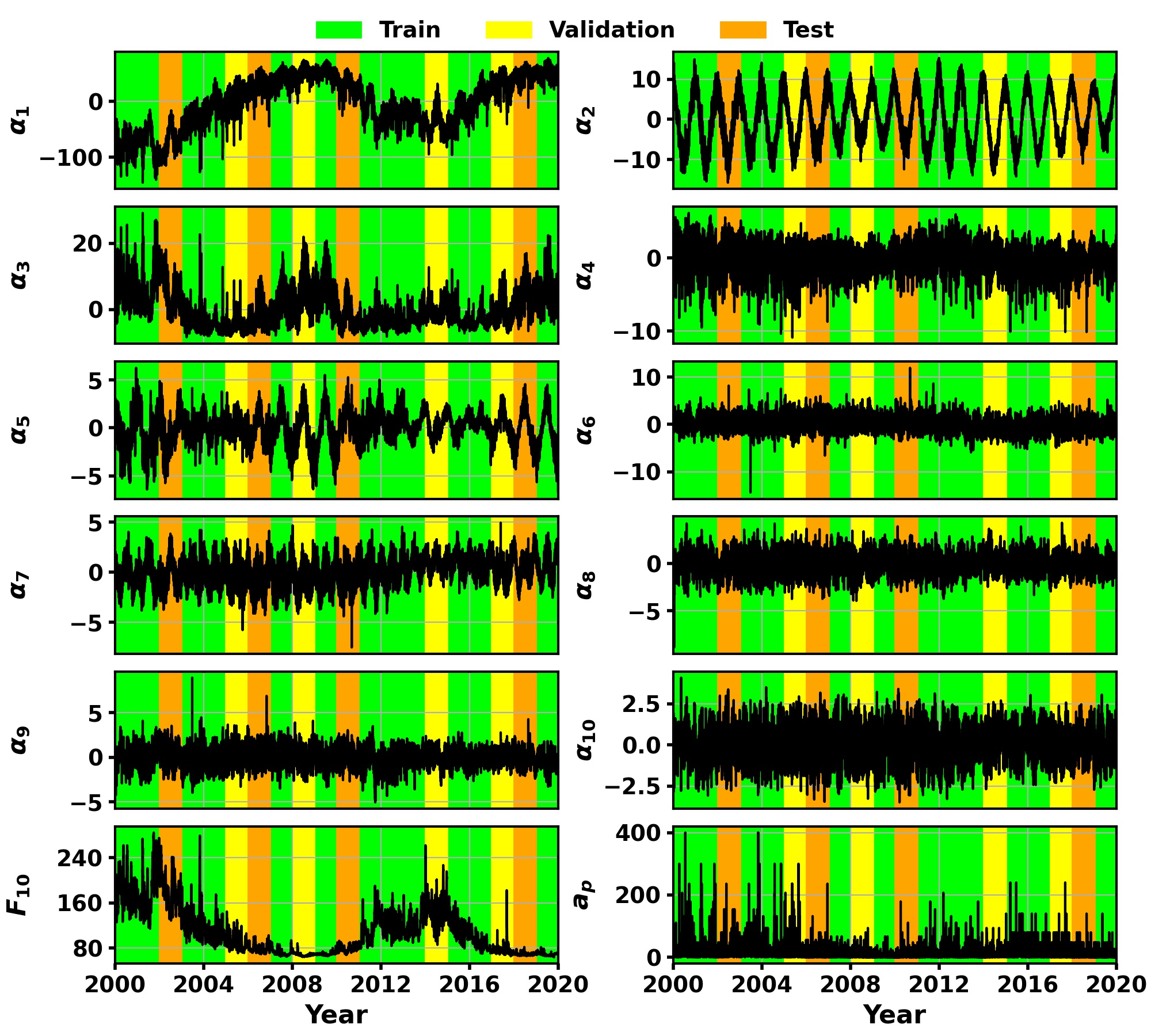}
	\caption{First 10 HASDM PCA coefficients with \textit{F\textsubscript{10}} and \textit{a\textsubscript{p}}. The shading represents the training, validation, and test sets.}
	\label{f:HASDM_PCA_TVT}
\end{figure}

The 58,440 samples in the HASDM dataset are split into 60\% training, 20\% validation, and 20\% test data. This is displayed in Figure \ref{f:HASDM_PCA_TVT}. As the number of training and validation samples is manageable, the full sets are used in tuning. We obtain the HASDM models directly from the tuner without a need to train further.

The CHAMP dataset is significantly larger with over 25 million total samples. Unlike the HASDM dataset, location is now an input. CHAMP only covers the local solar time domain once every three months. The dataset also does not span an entire solar cycle. Therefore, we repeat the following data split scheme. Eight weeks are used for training (483,840 samples), then the following week is used for validation (60,480 samples), and the next week is used for the test set (60,480 samples). This results in similar input and output distributions while keeping temporally disjoint sets as there are two weeks or 120,960 samples between the training segments. For the tuner, 1 million random samples are chosen from the training data and 500,000 random samples are chosen from the validation data. Once the tuner is complete, the best models are retrained on the full training set and evaluated on the other two sets.

\subsection{Uncertainty Quantification} \label{sec:UQ}

A common method for uncertainty quantification is the use of Gaussian process (GP) models. GP regression models are a supervised learning technique that can provide predictions with probability estimates in both regression and classification tasks \cite{GPR}. A GP model essentially provides a distribution of functions that fit the data which allows us to obtain a mean and variance for any prediction \cite{GPR_tutorial}. GP regression has been used recently in space weather applications in both driver forecasting \cite{GPR_Dst} and empirical model calibration \cite{GP_Cal}. Some limitations for GP implementations are difficult interpretation of results for multivariate problems and computational cost with large datasets \cite{GPR_DB}. The size of these datasets, particularly CHAMP, inhibits the use of GP modeling. Ensemble modeling -- the combination of predictions from multiple models -- is another common approach to obtain uncertainty estimates \cite{ensemble}. Limitations to ensemble methods include increased computational complexity and difficult optimization \cite{ensemble_challenge1,ensemble_challenge2}. We instead use two ML techniques: MC dropout (Section \ref{sec:MCDO}) and direct probability distribution prediction (Section \ref{sec:DPDP}), as UQ with machine-learned models is fairly unexplored in the space weather domain. Dropout is a generalization technique that applies Bernoulli distributions in each layer to change the flow of information through the model \cite{DO,DO2}. Dropout is traditionally only active during training to maintain a deterministic form in prediction. By forcing dropout to remain active in prediction, the model becomes probabilistic. MC dropout has been shown to be an approximation of a GP \cite{gal2016dropout}. For both methods, we use the negative logarithm of predictive density (NLPD) loss function (Equation \ref{eqNLPD} in Section \ref{sec:loss}). Licata et al. (2021) found that the mean square error loss function resulted in underestimated uncertainty estimates in surrogate modeling for the HASDM dataset \cite{HASDM_ML}. 

\subsubsection{Monte Carlo Dropout Implementation}\label{sec:MCDO}

The typical input and output shape is \textit{n} $\times$ \textit{n\textsubscript{inp}} and \textit{n} $\times$ \textit{n\textsubscript{out}}, respectively. \textit{n} is the number of samples, \textit{n\textsubscript{inp}} is the number of inputs, and \textit{n\textsubscript{out}} is the number of outputs. In training, the mean and standard deviation need to be unique to each input sample, so the model has to be provided each input \textit{k} times. \textit{k} needs to be a large enough number to allow for adequate representation of the predicted distribution. The inputs and outputs for training are stacked about a repeated intermediary axis. The training samples are identical about \textit{k}, but are unique about \textit{n}. The new input and output shapes -- necessary for proper training -- are \textit{n} $\times$ \textit{k} $\times$ \textit{n\textsubscript{inp}} and \textit{n} $\times$ \textit{k} $\times$ \textit{n\textsubscript{out}}, respectively. In each training batch, the mean and standard deviation are taken with respect to the intermediate axis, and the NLPD loss can be computed.

\subsubsection{Direct Probability Distribution Prediction}\label{sec:DPDP}

Another way to represent uncertainty is to directly predict the mean and standard deviation. The mean square error loss function cannot be used here as there are no labels for the standard deviation. However, Nix and Weigend (1994) used a neural network to directly predict the mean and variance of a toy dataset using the NLPD loss function \cite{Nix}. We implement this technique for the datasets presented. To accomplish this, we create a custom output layer with 2\textit{n\textsubscript{out}} neurons. The first \textit{n\textsubscript{out}} neurons represent the mean prediction and have a linear activation function. The last \textit{n\textsubscript{out}} neurons represent the standard deviation and use the softplus activation function. The softplus function and its derivative -- the sigmoid function -- are shown in Equation \ref{eqAct}.
%%%%%
\begin{equation} \label{eqAct}
\begin{split}
f(x) = ln(1+e^x) \;\;\;\;\;\;\;\;\;\;\; f'(x) = \frac{e^x}{1+e^x}
\end{split}
\end{equation}
%%%%%
The desired qualities of the standard deviation output are: (1) always positive and (2) having no upper bound. The initial choice was the absolute value function. However, the resulting models had erratic loss values, and it was difficult to obtain a good model. The softplus function is (1) always positive, (2) has no upper bound, (3) is monotonically increasing, and (4) is differentiable across all inputs. This resulted in stable training losses and better models.

\subsubsection{Metrics}

To compare the predictive capability of the models developed, we look at the mean absolute error across the training, validation, and test sets. The errors across different space weather conditions will be investigated as well. We also test the reliability of the uncertainty estimates both qualitatively and quantitatively. The calibration error score is given as
%%%
\begin{equation} \label{eq6}
\begin{split}
\textrm{Calibration Error} = \frac{100\%}{m \cdot n_{out}}\sum^{n_{out}}_{i=1} \sum^m_{j=1} \Big|p(z_{i,j})-p(\hat{z}_{i,j})\Big|
\end{split}
\end{equation}
%%%
where \textit{m} is the number of prediction intervals (PIs) of interest. Here the PIs range from 5\% to 95\% with 5\% increments in addition to 99\% -- [0.05, 0.10, 0.15, ... , 0.90, 0.95, 0.99]. $p(\hat{z}_{i,j})$ is the observed cumulative probability obtained by dividing the number of true samples within the prediction interval by the total number of samples. Equation \ref{eq6} is the miscalibration of prediction intervals averaged over each output and prediction interval tested. For this work, it provides the average deviation from all 20 PIs for each model output. We can visualize the reliability of the uncertainty estimates by plotting the calibration curve -- $p(\hat{z})$ vs $p(z)$. 

\section{Toy Problems}

\begin{figure}[htb!]
	\centering
	\small
	\includegraphics[width=0.90\textwidth]{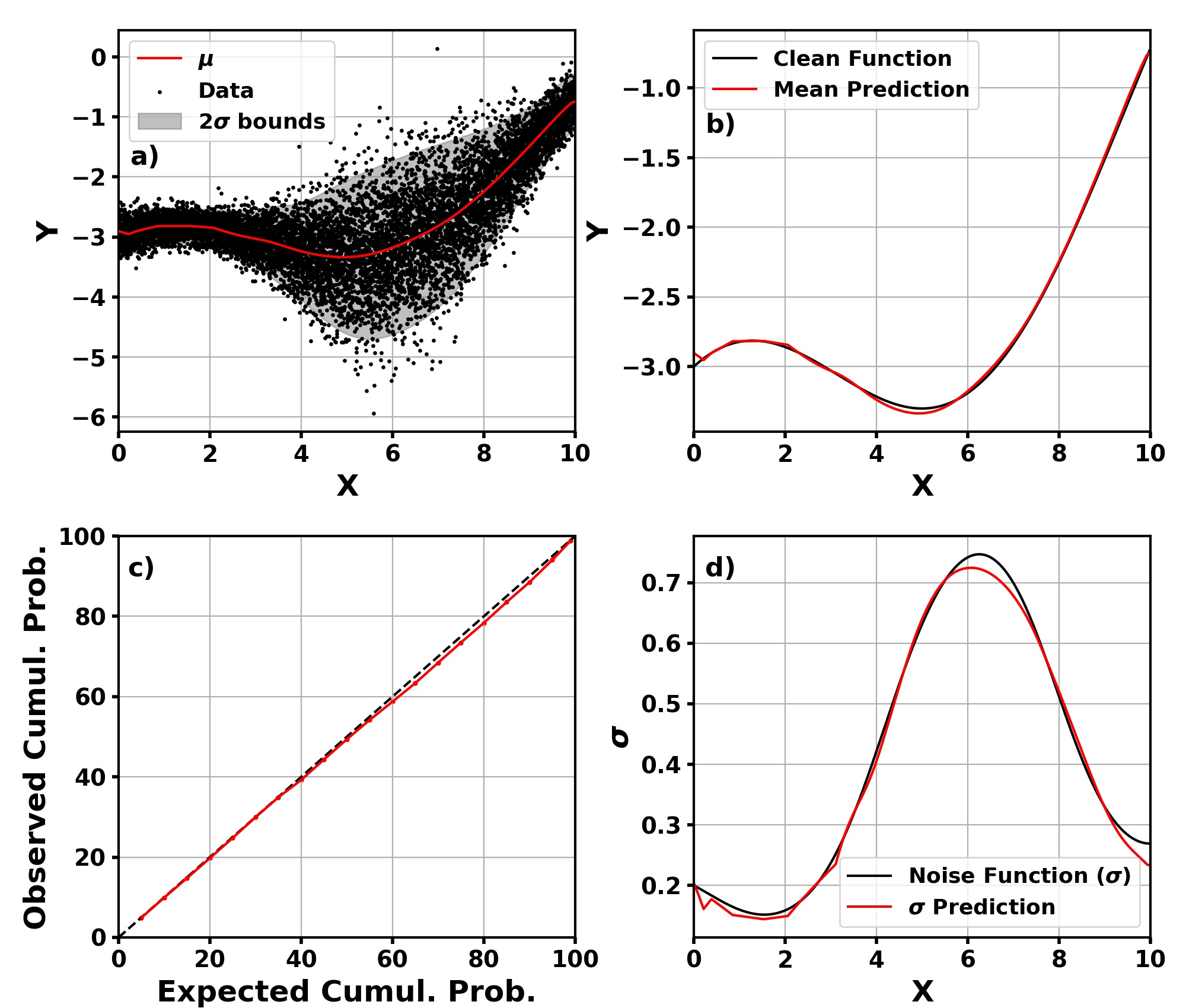}
	\caption{Mean prediction with $2\sigma$ bounds plotted on data (a), clean function plotted with mean prediction (b), calibration curve (c), and predicted standard deviation on true standard deviation function (d) for Problem 1.}
	\label{f:P1}
\end{figure}

% improve transition
To visualize the way the NLPD loss function influences training, we train models for two toy problems. Each problem is a function, $y(x)$, with additive Gaussian noise having zero-mean and a functional form to the standard deviation. These functions are displayed in Table \ref{t:functions}. The results for Problem 1 is shown in Figure \ref{f:P1} 

\vspace{0.5cm}

\begin{table}[htb!]
	\fontsize{10}{10}\selectfont
    \caption{Functions for the two toy problems with the right column being the functional form of the Gaussian noise.}
   \label{t:functions}
        \centering 
   \begin{tabular}{c | c | c } % Column formatting, 
      \hline 
          & \textbf{Function} & $\mathbf{\sigma}$ \\ \hline
         \textbf{Problem 1} & $0.3x + cos(0.5x) - 4 + \mathcal{N}(0,\sigma)$ & $0.5$\Large{$\frac{e^{sin(0.2x)}}{1+e^{sin(0.8x)}}$} \\ \hline
         \textbf{Problem 2} & $sin\left(2x+cos(3x)\right) + \mathcal{N}(0,\sigma)$ & $0.05sin(0.2x)$ \\ \hline
   \end{tabular}
\end{table}

\vspace{0.5cm}

\begin{figure}[htb!]
	\centering
	\small
	\includegraphics[width=0.90\textwidth]{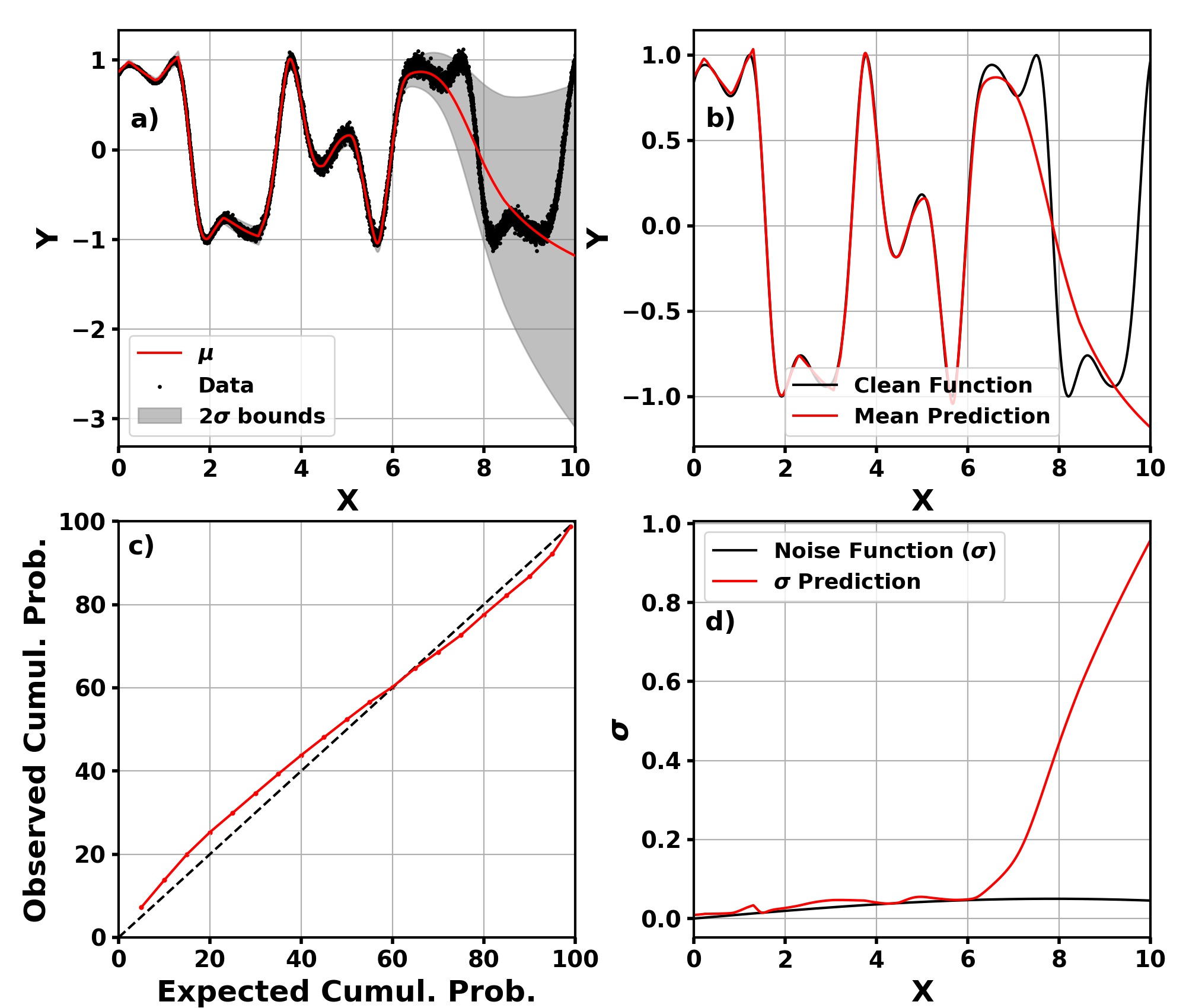}
    \caption{Mean prediction with $2\sigma$ bounds plotted on data (a), clean function plotted with mean prediction (b), calibration curve (c), and predicted standard deviation on true standard deviation function (d) for Problem 2.}
	\label{f:P2}
\end{figure}

Figure \ref{f:P1} shows that the model is able to adequately predict the function and is able to predict the overall probability distribution. The interesting aspect of the figure is panel (d): the model is able to predict the standard deviation without a label. Meanwhile, this is fairly trivial data. Figure \ref{f:P2} shows the predictions and calibration curve for the more complex Problem 2.

For the more complex data, the model is not as accurate over all $x$. When $x$ < 6, the model can accurately predict the mean and standard deviation. When $x$ > 6, the standard deviation prediction no longer represents uncertainty in the data but the model's uncertainty in its prediction. For this portion of panel (b), the mean prediction deviates from the true mean of the data and the standard deviation in panel (d) consequently increases. Panel (c) shows that the model is still well-calibrated and representing both uncertainty in the data and uncertainty in the model's predictions.

The NLPD loss function does not ensure model calibration. However, we show that it can be used -- if properly tested -- in model development to represent uncertainty in the data and uncertainty in the model's predictions. Note: these models were trained on the entire dataset, and this is purely for demonstration. The thermospheric density models are developed with separate validation and independent test sets.

\vspace{1cm}

\section{HASDM Model} \label{sec:HASDM_results}

Using the best tuner models for MC dropout and direct probability distribution prediction, we assess the error and calibration statistics. Table \ref{t:HASDM_stats} shows the mean absolute error and calibration error score for both techniques across the training, validation, and test sets.

\begin{table}[htb!]
	\fontsize{10}{10}\selectfont
    \caption{HASDM modeling results using MC dropout and direct probability prediction. Error refers to mean absolute error, and calibration is computed using Equation \ref{eq6}.}
   \label{t:HASDM_stats}
        \centering 
   \begin{tabular}{c | c | c | c} % Column formatting, 
      \hline 
         \textbf{Metric} & \textbf{Set} & \textbf{MC Dropout} & \textbf{Direct Probability} \\ \hline
         \multirow{3}{*}{\textbf{Error}} & \textbf{Training} & 9.07\% & \textbf{8.55\%} \\ 
         & \textbf{Validation} & 10.69\% & \textbf{9.91\%} \\
         & \textbf{Test} & 10.69\% & \textbf{10.60\%} \\ \hline
         \multirow{3}{*}{\textbf{Calibration}} & \textbf{Training} & 3.06\% & \textbf{1.74\%} \\
         & \textbf{Validation} & 2.51\% & \textbf{2.45\%} \\
         & \textbf{Test} & \textbf{1.76\%} & 2.81\% \\ \hline
   \end{tabular}
\end{table}

\begin{figure}[htb!]
	\centering
	\small
	\includegraphics[width=0.92\textwidth]{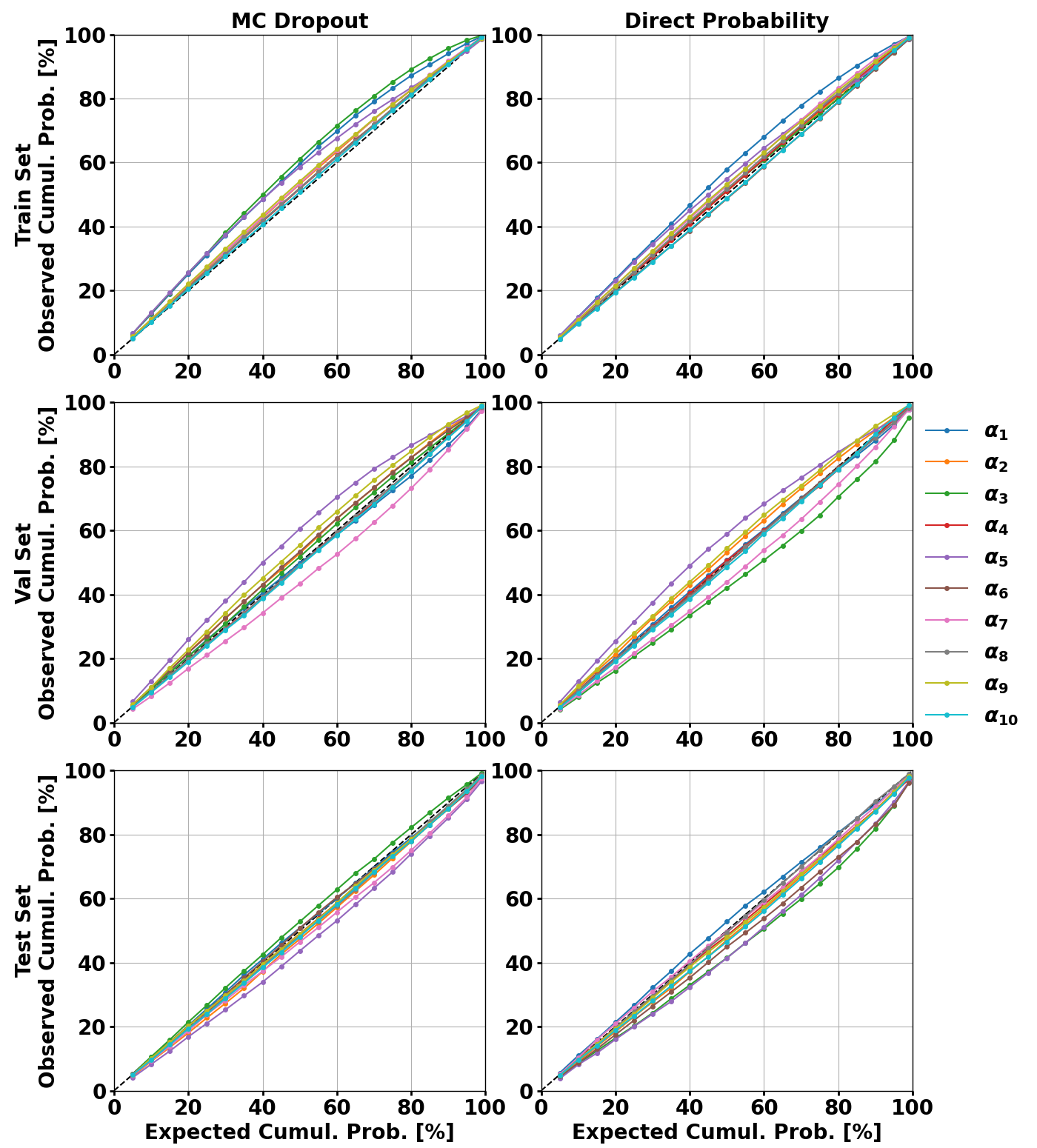}
    \caption{The left and right columns show the MC dropout and direct probability calibration curves, respectively. The top, middle, and bottom rows are the calibration curves for the training, validation, and test sets, respectively.}
	\label{f:HASDM_cal}
\end{figure}

It is evident that the performance using both methods is very similar. Across all three sets, the mean absolute error and calibration error score do not deviate by more than 0.8\% and 1.4\% respectively. The MC dropout model has better performance on the independent test set in terms of calibration. This is a desired quality as the test data is not used for model development in any way. As the calibration error scores are composites of the scores for each output, the calibration curves are shown in Figure \ref{f:HASDM_cal} for a qualitative assessment.

Both techniques lead to slightly overestimated uncertainties on the training set for multiple outputs. Meanwhile, the remaining outputs are almost perfectly calibrated. On the validation set, each model has outputs with overestimated and underestimated uncertainties. Again, most of the outputs are very well-calibrated which is affirmed by the calibration error scores. For the test set, the direct probability prediction model tends to marginally underestimate the uncertainty while the MC dropout model provides reliable uncertainty estimates on virtually all model outputs. Table \ref{t:HASDM_conditions} shows the mean absolute error for both models across an array of solar and geomagnetic conditions. The entire dataset is used for this analysis as there are not enough samples in each bin using only the test set.

\vspace{5mm}

\begin{table}[htb!]
	\fontsize{10}{10}\selectfont
    \caption{Mean absolute error across global grid for HASDM-ML as a function of space weather conditions.}
   \label{t:HASDM_conditions}
        \centering 
   \begin{tabular}{c | c | c | c | c | c} % Column formatting, 
      \hline 
      \multicolumn{6}{c}{\textbf{MC Dropout}} \\ \hline
          & $\mathbf{F_{10} \leq 75}$ & $\mathbf{75 < F_{10} \leq 150}$ & $\mathbf{150 < F_{10} \leq 190}$ & $\mathbf{F_{10} > 190}$ & \textbf{All} $\mathbf{F_{10}}$\\ \hline
         $\mathbf{a_p \leq 10}$ & 8.96\% & 9.78\% & 9.97\% & 9.14\% & 9.50\% \\ \hline
         $\mathbf{10 < a_p \leq 50}$ & 9.76\% & 10.05\%  & 10.87\%  & 9.90\% & 10.09\% \\ \hline
         $\mathbf{a_p > 50}$ & 15.35\% & 12.86\% & 13.23\%  & 12.55\% & 13.01\% \\ \hline
         \textbf{All} $\mathbf{a_p}$ & 9.12\% & 9.92\% & 10.36\% & 9.55\% & 9.71\% \\ \hline
      \multicolumn{6}{c}{\textbf{Direct Probability}} \\ \hline
          & $\mathbf{F_{10} \leq 75}$ & $\mathbf{75 < F_{10} \leq 150}$ & $\mathbf{150 < F_{10} \leq 190}$ & $\mathbf{F_{10} > 190}$ & \textbf{All} $\mathbf{F_{10}}$\\ \hline
         $\mathbf{a_p \leq 10}$ & 8.64\% & 9.33\% & 9.35\% & 9.11\% & 9.10\% \\ \hline
         $\mathbf{10 < a_p \leq 50}$ & 9.18\% & 9.51\%  & 9.69\%  & 9.64\% & 9.48\% \\ \hline
         $\mathbf{a_p > 50}$ & 11.14\% & 11.23\% & 11.34\%  & 10.30\% & 11.11\% \\ \hline
         \textbf{All} $\mathbf{a_p}$ & 8.74\% & 9.42\% & 9.52\% & 9.34\% & 9.23\% \\ \hline
   \end{tabular}
\end{table}

\vspace{5mm}

These errors tend to reiterate the results from Table \ref{t:HASDM_stats}. The direct probability model was more accurate on all three sets, and Table \ref{t:HASDM_conditions} shows that it is also more accurate across all 20 conditions considered. For a majority of the conditions, the difference is small (< 1\%). However, the high \textit{a\textsubscript{p}} conditions show that the direct probability model makes considerable improvements. These error reduction from MC dropout range from 1.6 -- 4.1\%.

\begin{figure}[htb!]
	\centering
	\small
	\includegraphics[width=0.60\textwidth]{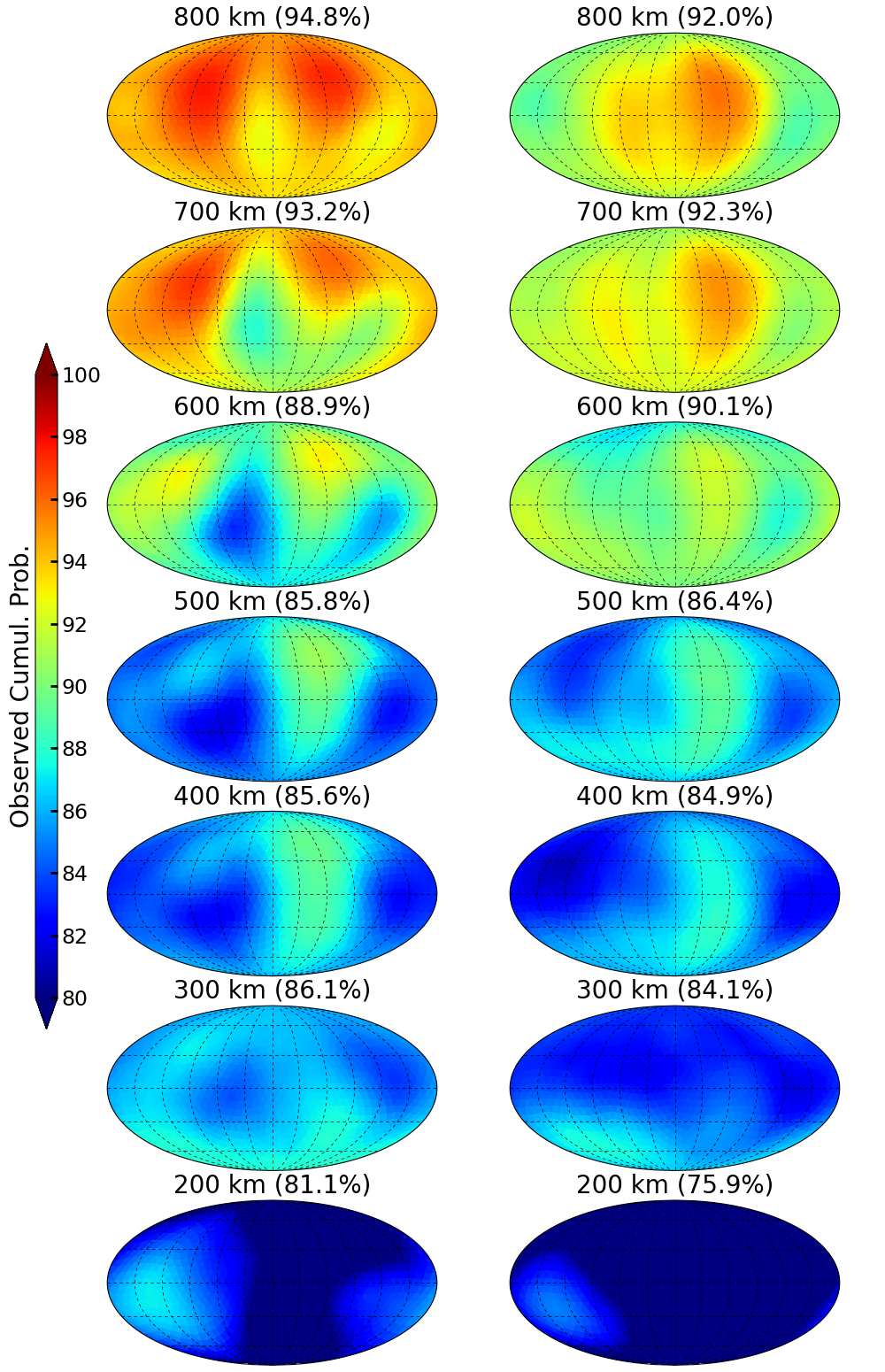}
    \caption{Observed cumulative probability maps for a 90\% prediction interval using the MC dropout (left) and direct probability (right) models. The average observed cumulative probability is shown for each altitude in parenthesis.}
	\label{f:HASDM_Maps}
\end{figure}

To further assess the uncertainty capabilities of the models, we attempt to visualize the calibration in the full-state (global density grids) to identify any spatial dependence in the reliability of the uncertainty estimates. First, the models are evaluated on the entire test set and the density mean and standard deviations are extracted. Using these statistics, the observed cumulative probability with a 90\% prediction interval is computed for each spatial location. The resulting $24\times19\times27$ array is used to determine how well calibrated the model is on independent data as a function of location. We show seven maps for each model (200, 300, ... , 800 km) in Figure \ref{f:HASDM_Maps}. Even though HASDM has a lateral spatial resolution of 24 longitude and 19 latitude segments, we interpolate the results to the polyhedral grid used in the EXTEMPLAR model for visualization purposes. This is done in the remainder of the manuscript.

For reference, perfect calibration in Figure \ref{f:HASDM_Maps} would be uniform green maps at all altitudes. This would convey that with a 90\% prediction interval, the model's predictions/uncertainty estimates contain 90\% of true samples at all locations. While this is not the case, the results are still insightful. At 200 km, both models are underestimating the uncertainty by 10 -- 15\%. This could be a result of the relative variability as a function of altitude in the SET HASDM density database. The general trend of relative variability is that it increases with altitude, so the models may underpredict the standard deviation at low altitudes as a result, which indicates that the model has a false sense of confidence in that region. Both models have an average cumulative probability within 5\% of the expected value at most of the altitudes shown in Figure \ref{f:HASDM_Maps} with the best results at 600 km. At 700 and 800 km, both models begin to overestimate uncertainty, likely because they have the lowest confidence at those altitudes. An interesting outcome of this study is the lateral variability of the cumulative probability between the models. The MC dropout model (left) has more lateral variability, meaning the cumulative probability changes more as a function of longitude and latitude.

\clearpage

\section{CHAMP Model}

After running tuners for both uncertainty techniques, we trained the best models on the entire training set. The models were chosen based on the lowest prediction error and best calibration scores on the validation set. Table \ref{t:CHAMP_stats} shows the mean absolute error and calibration error scores on the three sets.

\begin{table}[htb!]
	\fontsize{10}{10}\selectfont
    \caption{CHAMP modeling results using MC dropout and direct probability prediction. Error refers to mean absolute error, and calibration is computed using Equation \ref{eq6}.}
   \label{t:CHAMP_stats}
        \centering 
   \begin{tabular}{c | c | c | c} % Column formatting, 
      \hline 
         \textbf{Metric} & \textbf{Set} & \textbf{MC Dropout} & \textbf{Direct Probability} \\ \hline
         \multirow{3}{*}{\textbf{Error}} & \textbf{Training} & 13.13\% & \textbf{12.59\%} \\ 
         & \textbf{Validation} & 13.67\% & \textbf{12.82\%} \\
         & \textbf{Test} & 13.14\% & \textbf{12.62\%} \\ \hline
         \multirow{3}{*}{\textbf{Calibration}} & \textbf{Training} & \textbf{3.93\%} & 5.84\% \\
         & \textbf{Validation} & 0.64\% & \textbf{0.25\%} \\
         & \textbf{Test} & \textbf{0.22\%} & 0.37\% \\ \hline
   \end{tabular}
\end{table}

\begin{figure}[htb!]
	\centering
	\small
	\includegraphics[width=0.7\textwidth]{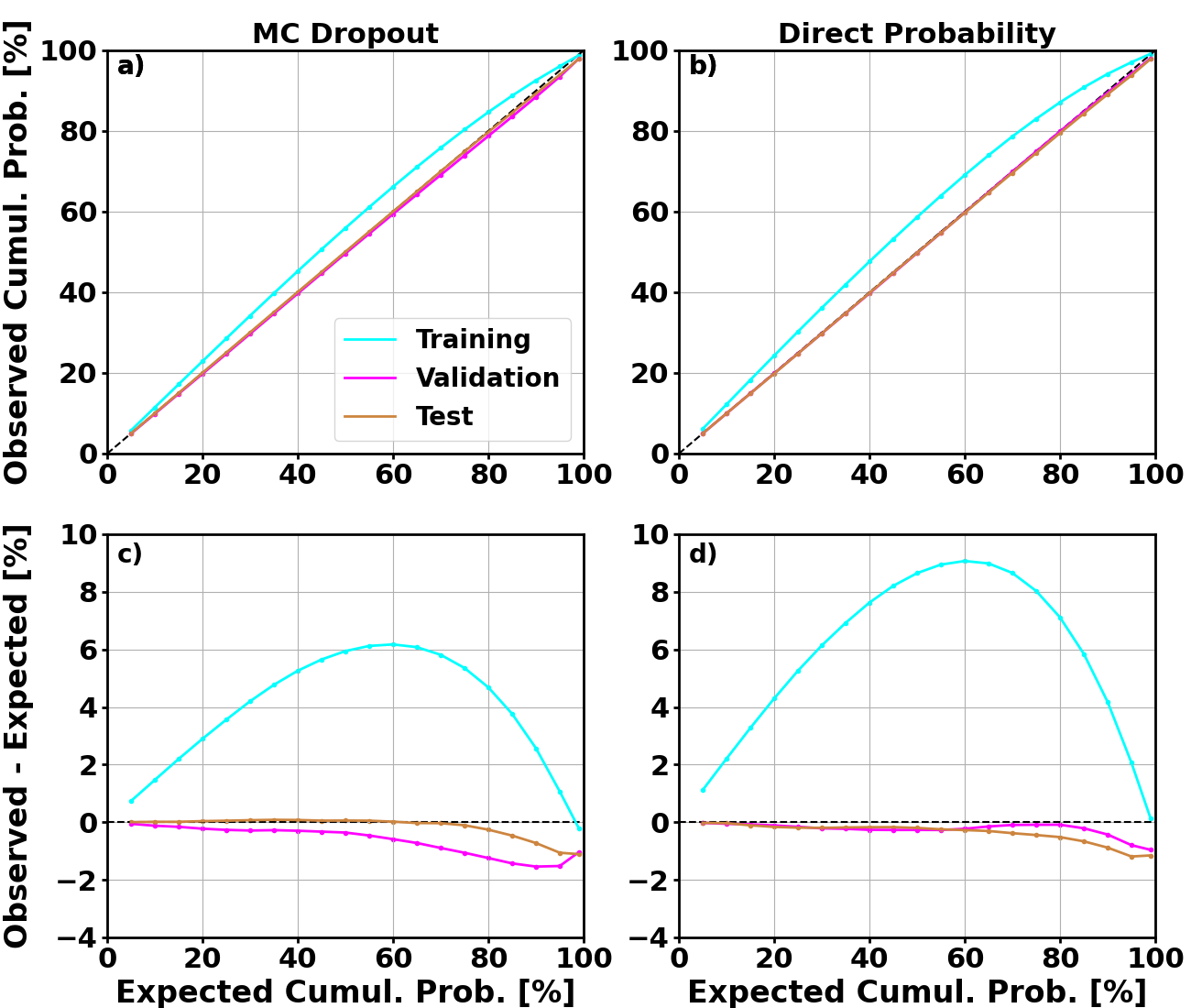}
    \caption{Calibration curves for the training, validation, and test sets using MC dropout (a) and direct probability prediction (b). Panels (c) and (d) show the difference between the observed and expected cumulative probability using MC dropout and direct probability prediction, respectively.}
	\label{f:CHAMP_cal}
\end{figure}

Both models are well-generalized in terms of prediction accuracy. The range in error between sets for the MC dropout and direct probability model is 0.54\% and 0.23\%, respectively. Both models have higher calibration error scores on the training set but have similar scores on the validation and test sets. The two techniques provide similar results with the only notable difference is the 1.91\% higher calibration error score for the direct probability model on the training set. The calibration curves for both models are shown in Figure \ref{f:CHAMP_cal}.

Both models are well-calibrated on all three sets. There is a tendency for both models to slightly overestimate uncertainty on the training set which is more evident for the MC dropout model. The differences between the calibration curves and the perfectly calibrated reference line (in black) is shown in panels (c) and (d). Panel (d) highlights the overestimation of uncertainty for the direct probability model on the training set. However, it never deviates by more than 9\%. Both models tend to underestimate uncertainty on the validation and test set for the larger prediction intervals. Again, the deviation from perfect calibration is no more than 2\% for any PI. Due to the intrinsic difference between the datasets that the CHAMP and HASDM models are developed from, the proceeding analyses will be different than those in Section \ref{sec:HASDM_results}.

\subsection{Global Modeling with Local Measurements}

\begin{figure}[htb!]
	\centering
	\small
	\includegraphics[width=1\textwidth]{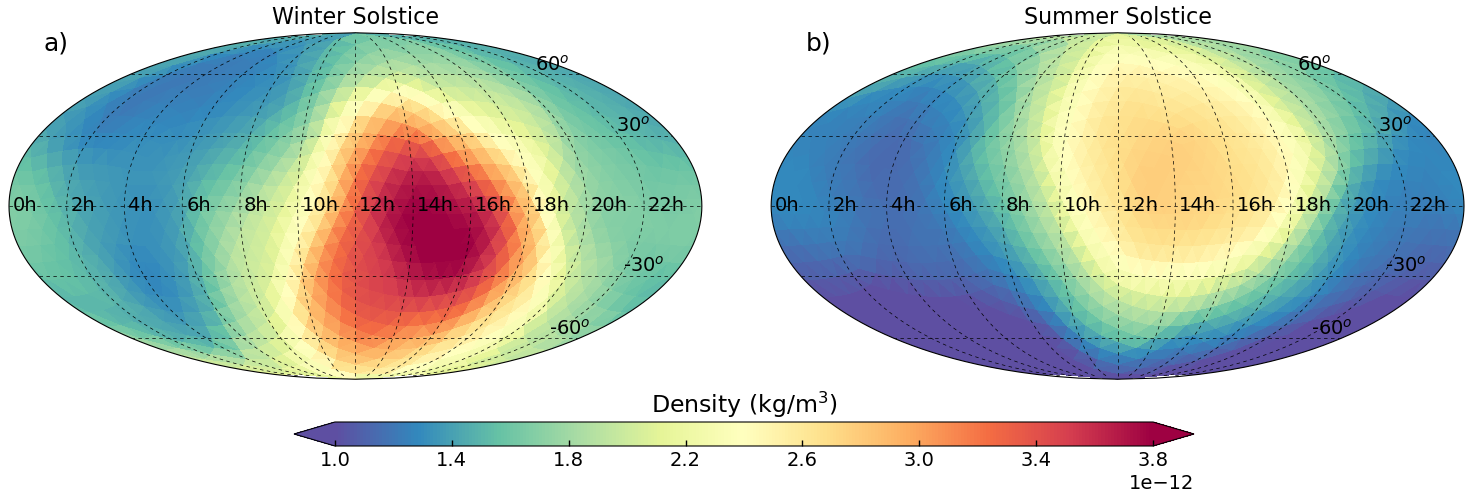}
    \caption{Global density map with moderate solar activity, low geomagnetic activity, the altitude fixed to 400 km, and the time of day being 00:00 UTC for the winter solstice (a) and the summer solstice (b).}
	\label{f:soltice}
\end{figure}

The CHAMP models were developed with in-situ measurements, but we hypothesize that it should be able to learn the functional relationship of the combined inputs. Therefore, the model should be able to provide global outputs at any point in time. As a qualitative assessment, we show global maps at 400 km for the winter and summer solstices in Figure \ref{f:soltice} using the direct probability model. All proceeding global analyses will be performed using this model. For this test, the solar drivers are all set to 120 sfu, \textit{SYM-H} is set to 0 nT, both Poynting flux totals are set to 27GW, and the time is set to 00:00 UTC.

The diurnal structure is present in both panels with the peak density being in the southern hemisphere during the winter solstice and in the northern hemisphere during the summer solstice. This shows the model's understanding on annual trends (Earth's tilt). The general density level is higher during the winter solstice, but the relative variation between day and night are very similar. This is reaffirmed by the exospheric temperature distribution shown by Weimer et al (2020) during the solstices. Additional global density maps at different altitudes can be found in Appendix \hyperref[sec:appA]{A} using baseline conditions. Furthermore, a global storm example is shown in Appendix \hyperref[sec:appB]{B}.

\begin{figure}[htb!]
	\centering
	\small
	\includegraphics[width=0.70\textwidth]{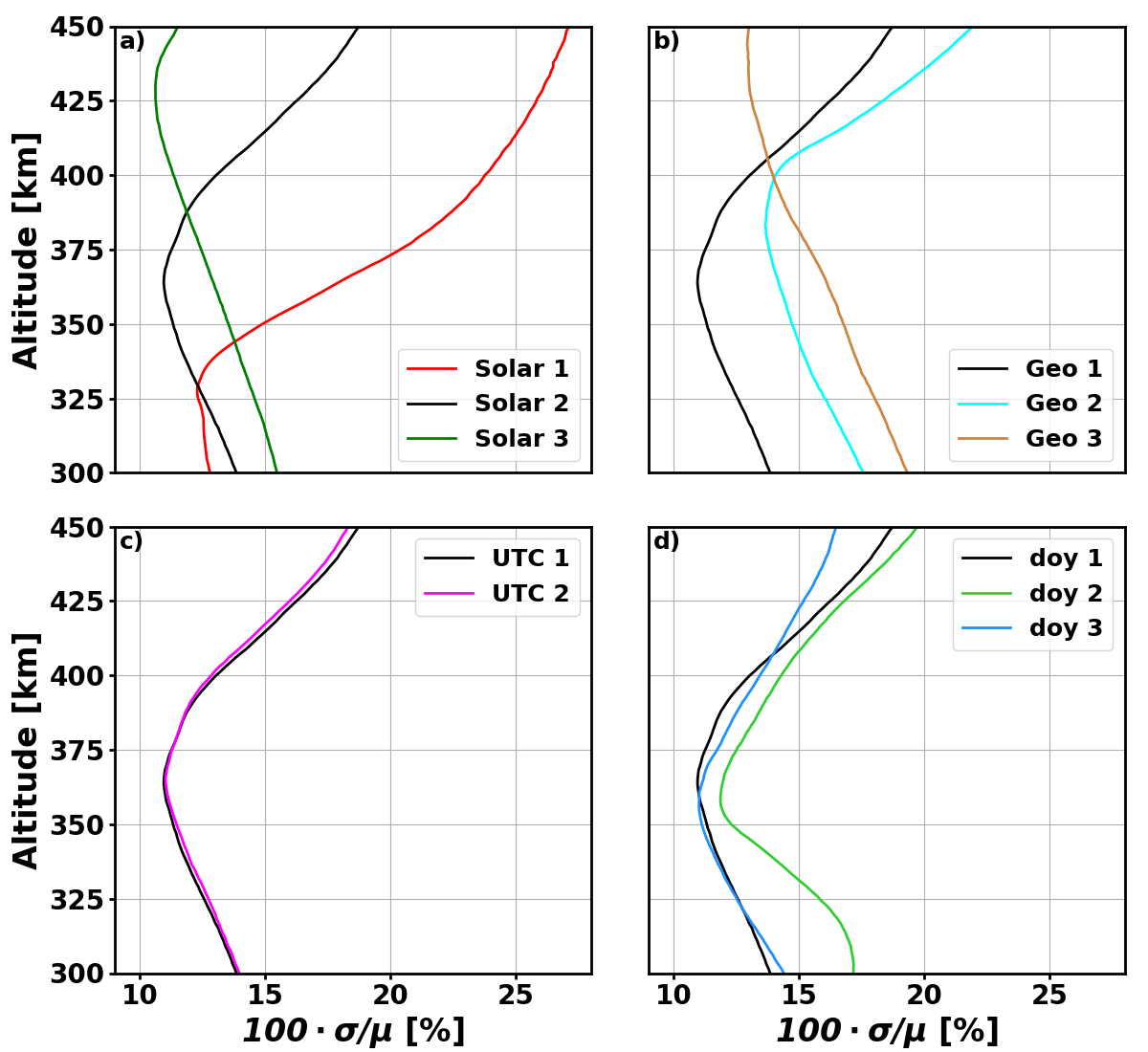}
    \caption{Normalized uncertainty variations as a function of altitude for solar (a), geomagnetic (b), daily (c), and annual (d) cases. The drivers for each curve can be found in Table \ref{t:CHAMP_conditions}.}
	\label{f:alt_err}
\end{figure}

Next, we look at the uncertainty levels for eight unique conditions of activity and time. These are all displayed in Table \ref{t:CHAMP_conditions}. Using these space weather and temporal inputs, the CHAMP model is evaluated at all 1,620 polyhedral grid locations from 300 to 450 km in 1 km increments. The metric we use here is a normalized measure of model uncertainty: $100 \cdot \sigma / \mu$, essentially providing the 1-$\sigma$ uncertainty as a percentage of the mean prediction. The resulting maps are averaged across each altitude to evaluate the model's uncertainty for each condition as a function of altitude. Three aspects of model drivers are investigated: solar activity, geomagnetic activity, and temporal dependence. In Table \ref{t:CHAMP_conditions}, there are three solar activity levels, with all other drivers kept constant. There are also three geomagnetic cases: low and high geomagnetic activity with moderate solar activity, and high geomagnetic activity with high solar activity. We only look at two daily cases -- 00:00 and 12:00 UTC. We also look at the fall equinox, summer solstice, and winter solstice with moderate solar and low geomagnetic activity. The resulting altitude profiles are shown in Figure \ref{f:alt_err}.

\begin{table}[htb!]
	\fontsize{10}{10}\selectfont
    \caption{CHAMP model inputs to study various conditions as a function of altitude. * Solar 2 is also considered Geo 1, UTC 1, and doy 1.}
   \label{t:CHAMP_conditions}
        \centering 
   \begin{tabular}{c | c | c | c | c | c } % Column formatting, 
      \hline 
          & \textbf{Solar Drivers} & \multicolumn{2}{c|}{\textbf{Geomagnetic Drivers}} & \multicolumn{2}{c}{\textbf{Temporal Drivers}} \\ \hline
         \textbf{Condition Name} & \textbf{FMSY} & \textbf{SYM-H} & \textbf{S\textsubscript{N} = S\textsubscript{S}} & \textbf{UTC} & \textbf{doy} \\ \hline
         \textbf{Solar 1} & 75 & 0 & 27 & 0 & 262 \\ \hline
         \textbf{Solar 2*} & 120 & 0 & 27 & 0 & 262 \\ \hline
         \textbf{Solar 3} & 190 & 0 & 27 & 0 & 262 \\ \hline
         \textbf{Geo 2} & 120 & -75 & 128 & 0 & 262 \\ \hline
         \textbf{Geo 3} & 190 & -75 & 128 & 0 & 262 \\ \hline
         \textbf{UTC 2} & 120 & 0 & 27 & 12 & 262 \\ \hline
         \textbf{doy 2} & 120 & 0 & 27 & 0 & 172 \\ \hline
         \textbf{doy 3} & 120 & 0 & 27 & 0 & 355 \\ \hline
   \end{tabular}
\end{table}

Panel (a) in Figure \ref{f:alt_err} shows that the CHAMP model has low uncertainty in its lower altitude predictions for solar minimum (or low solar activity) which drastically increases with altitude. The opposite can be said for solar maximum. The moderate solar activity case results in lowest uncertainties between 350 and 375 km and higher uncertainties above and below that range. This is all a result of CHAMP's altitude from 2002-2010. It started around 460 km during solar maximum and ended at 300 km during solar minimum. Therefore, the model has confident predictions in the altitude range the satellite was located during the various phases of the solar cycle. If there was additional data from satellites at different altitudes over a longer time period, the model would likely be more confident over a larger altitude range.

In panel (b), we see the same general trends for Geo 1 and Geo 2, because they are evaluated using moderate solar activity. However, it is evident that the increase in geomagnetic activity results in up to 5\% more uncertainty. The Geo 3 case is similar to Solar 3 (high solar activity) but again has increased uncertainty due to the storm conditions it represents. Panel (c) indicates that there is a low impact from universal time on the model uncertainty. In Panel (d), the black line indicates the fall equinox which is similar to the winter solstice. The Winter solstice uncertainties deviate from the equinox uncertainties at the highest altitude range. While the overall shape remains consistent, there are highest uncertainties for the summer solstice at all altitudes. The overall takeaway form Figure \ref{f:alt_err} is that the shape of the model uncertainty altitude profile is most strongly effected by the solar activity level while the day of year and geomagnetic activity tend to uniformly increase or decrease uncertainty. These profiles would all likely be impacted if the model was developed using additional satellite data.

\section{Evaluation Time Comparison}

We attempt to provide an equal comparison of the two methods in terms of computational complexity. To do so, each CHAMP model is evaluated on either 8,640 samples (one week) or 86,400 samples (ten weeks). For the direct probability prediction model, it sees each input once and provides the mean and standard deviation. These are used to sample a Gaussian distribution 1,000 times to get probabilistic predictions for density over the given window.

For MC dropout, we cannot pass one week of inputs to the model stacked 1,000 times (as is done for HASDM). There is not enough memory on an NVIDIA GeForce RTX 2080 Ti graphics processing unit (GPU) -- 11 GB -- to perform this evaluation. Therefore, we pass the 100 repeated inputs in 10 chunks to obtain the 1,000 predictions. When evaluating over ten weeks, we must reduce to 10 repeated inputs in 100 chunks. In Table \ref{t:runtime}, we show the evaluation times on both GPU and CPU for both methods over the two durations. Note: when running MC dropout on CPU, we use 100 repeated inputs for both durations. The batch size for all predictions is 2\textsuperscript{17} or 131,072. The size of the MC dropout and direct probability models are 233.3 kB and 21.9 MB, respectively.

\begin{table}[htb!]
	\fontsize{10}{10}\selectfont
    \caption{Run time to obtain 1,000 probabilistic predictions from each model using GPU and CPU in seconds.}
   \label{t:runtime}
        \centering 
   \begin{tabular}{c | c | c | c} % Column formatting, 
      \hline 
         \textbf{Method} & \textbf{Samples} & \textbf{GPU Run Time} & \textbf{CPU Run Time} \\ \hline
         \multirow{2}{*}{\textbf{MC Dropout}} & 8,640 & 2.11 & 13.65 \\ 
         & 86,400 & 18.29 & 127.79 \\ \hline
         \multirow{2}{*}{\textbf{Direct Probability}} & 8,640 & 0.58 & 0.52 \\
         & 86,400 & 3.93 & 3.93 \\ \hline
   \end{tabular}
\end{table}

The run times are unique to these specific models. The size of the models plays a role in run time, and the size of these models are a result of the tuner. The MC dropout model is approximately 100 times smaller, but the increase in required model prediction calls results in the significantly longer run times. The direct probability method, for this particular problem, is anywhere from 3 to 30 times faster depending on the number of samples and whether the GPU or CPU is being used.

\section{Discussion and Conclusions}

In this work, we leverage the NLPD loss function to develop thermospheric density models using 1) MC dropout and 2) direct probability prediction. These two uncertainty techniques were used to create both a model based in the PCA coefficients of the SET HASDM density database and a model based in localized accelerometer-derived density estimates from CHAMP. Using two toy problems, we showed that the NLPD loss function can be used to create a ML model with calibrated uncertainty estimates relative to uncertainty in the model, uncertainty in the data, or both. For the HASDM database, the MC dropout and direct probability distribution prediction models had similar metrics in terms of error and calibration. Furthermore, the calibration curves for the PCA coefficients were nearly identical. By looking at the density calibration of both HASDM models, we found that they were well-calibrated at mid-altitudes, and there was more lateral variability in the calibration of the MC dropout model.

The CHAMP models also had similar performance and were both well-generalized. We test the CHAMP model's global prediction capabilities by generating baseline maps during the winter and summer solstices to ensure physical global trends are being captured by the CHAMP model. This showed that the model was able to emulate the effect of Earth's tilt. We also performed global evaluations for eight unique conditions to determine the altitude dependence of model uncertainty. The altitude profiles showed that the minimum and maximum 1-$\sigma$ uncertainties were 10 -- 28\% of the mean predictions, respectively. Solar activity was most influential in determining the profiles' shapes while geomagnetic activity and day of year tended to provide uniform changes in the uncertainty. In general, the MC dropout and direct probability methods were shown to have similar performance for thermospheric density modeling applications. However, there are pros and cons for both methods, and careful consideration is required when deciding on a UQ method for space weather models. These are highlighted in Table \ref{t:pros_cons}.

\begin{table}[htb!]
	\fontsize{10}{10}\selectfont
    \caption{Pros and cons for MC dropout and direct probability distribution prediction.}
  \label{t:pros_cons}
        \centering 
  \begin{tabular}{c | l | l} % Column formatting, 
      \hline 
      \textbf{Method} & \multicolumn{1}{c|}{\textbf{Pros}} & \multicolumn{1}{c}{\textbf{Cons}} \\ \hline
      \multirow{2}{*}{\textbf{MC Dropout}} & \textbullet \; No need to sample from a Gaussian & \textbullet \; Longer evaluation times \\ 
      & \;\;\;distribution & \textbullet \; Not compatible with large datasets \\ \hline
      \multirow{2}{*}{\textbf{Direct Probability}} & \textbullet \; Only need single evaluation & \textbullet \; Required sampling from a Gaussian distribution \\ 
      & \textbullet \; Computationally efficient & \;\;\;to obtain probabilistic predictions \\ \hline
  \end{tabular}
\end{table}

The main disadvantage for the direct probability method is the requirement to sample from a Gaussian distribution to get probabilistic density variations. The drawback to MC dropout is its higher computational cost. In terms of density modeling, both techniques have prompt evaluation times. Relative to one another, we show that the direct probability models can be evaluated much faster. The size of the training data (in both number of samples and dimensionality) is also important to consider. With MC dropout, GPU memory can constrain modeling efforts if the dataset is too large. It can also require additional steps for prediction. In this work, MC dropout did not inhibit model development for the smaller HASDM PCA data. However, it did add numerous considerations when developing and evaluating the CHAMP model. In general, the uncertainty estimation capabilities may be improved through modifications to the loss function to either: a) add higher order moments or b) obtain non-Gaussian estimates.

All the preceding results show that for thermospheric density applications, these two techniques can be used to obtain an accurate model with reliable uncertainty estimates. There are other methods that can be used in space weather application such as GP regression and ensemble modeling, as previously mentioned, but this is a sufficient starting point. Other final considerations concern orthogonality and applicability. For a multi output regression model (e.g. HASDM models), the outputs must be orthogonal. This is to both prevent collinearity and since the use of NLPD requires uncorrelated outputs. The CHAMP data only spans an altitude range of 300 -- 460 km. Any predictions outside of this range may be unreliable. To combat this, density estimates from other satellites can be added to increase the altitude coverage and provide the model with more data to learn from, as discussed in Section \ref{sec:CHAMP}. 

\section*{Data Statement}

Requests can be submitted for access to the SET HASDM density database at \url{https://spacewx.com/hasdm/} and all reasonable requests for scientific research will be accepted as explained in the rules of road document on the website. The historical space weather indices used in this study can also be found at \url{https://spacewx.com/jb2008/} in the SOLFSMY.TXT, 	SOLRESAP.TXT, and DSTFILE.TXT files for the solar indices, \textit{ap}, and \textit{Dst}, respectively. Free and one-time only registration is required to access these files. \textit{SYM-H} data was obtained from \url{http://wdc.kugi.kyoto-u.ac.jp/aeasy/index.html} thanks to the World Data Center for Geomagnetism in Kyoto. CHAMP density estimates from Mehta et al. (2017) can be found at \url{http://tinyurl.com/densitysets}.

\section*{Acknowledgements}

PMM gratefully acknowledges support under NSF CAREER award \#2140204. The authors would like to acknowledge DLR for their work on the CHAMP mission along with GFZ Potsdam for managing the data.

\bibliographystyle{ieeetr}  
\bibliography{references}  %%% Remove comment to use the external .bib file (using bibtex).

\newpage

\section*{Appendix A: Global CHAMP Model Maps as a Function of Altitude}\label{sec:appA}

\begin{figure}[htb!]
	\centering
	\small
	\includegraphics[width=0.95\textwidth]{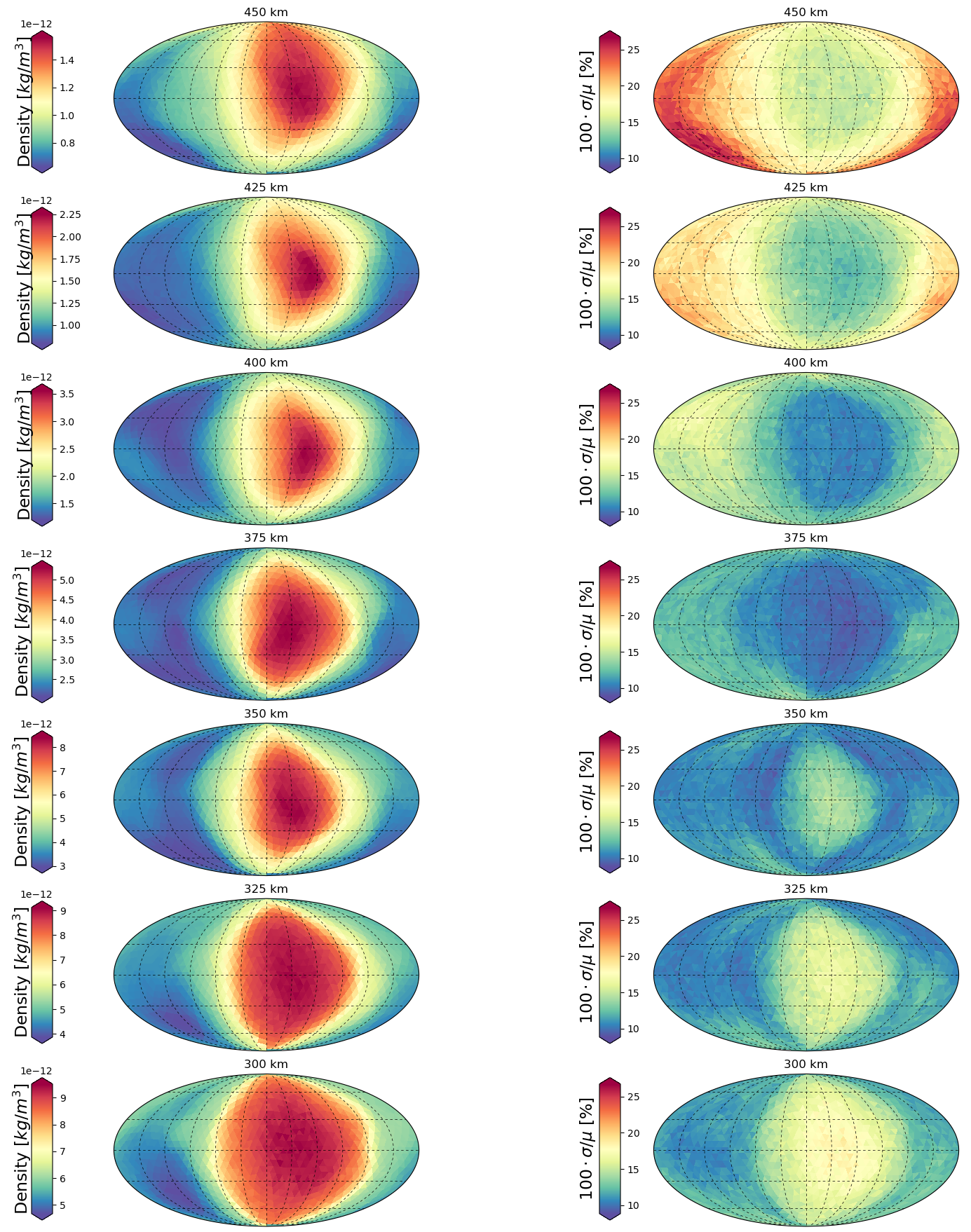}
    \caption{Global density maps (left) for moderate solar and low geomagnetic activity with the associated uncertainties (right). Note: there are different colorbar limits for the density maps but a consistent colorbar is used for all uncertainty maps.}
	\label{f:Altitude_Maps_CHAMP}
\end{figure}

To demonstrate global prediction of mean and standard deviation, the direct probability CHAMP model is evaluated at all EXTEMPLAR grid locations (for display purposes) for baseline conditions at seven altitudes. Figure \ref{f:Altitude_Maps_CHAMP} shows the global density maps and the associated uncertainties ($100 \cdot \sigma/\mu$). These maps show that the model makes sensible predictions at most altitudes. At 300 and 325 km, the model seems to predict a significant density maximum on the dayside that spans from 0800 to 2000 hours local time. Concurrently, the normalized uncertainty is largest in this region. Between 350 and 425 km, the predictions look as expected with an interesting trend of a shrinking region of high density on the dayside. The normalized uncertainties transition from most uncertain in the dayside to most uncertain in the night side. At 450 km, the peak shifts north and the uncertainties are on the order of 25\% of the mean prediction on the night side. This northern density shift is abnormal as these maps are all for the fall equinox. In general, the CHAMP model predicts normal density distributions between 350 and 425 km, where a majority of the training data resides. Relating these results to Figure \ref{f:alt_err}, the global density maps may all look more physical if the solar activity used for each altitude were more representative of the solar activity encountered by CHAMP at each altitude.

\clearpage

\section*{Appendix B: Storm Time Evaluation with Global Maps}\label{sec:appB}

\begin{figure}[htb!]
	\centering
	\small
	\includegraphics[width=1\textwidth]{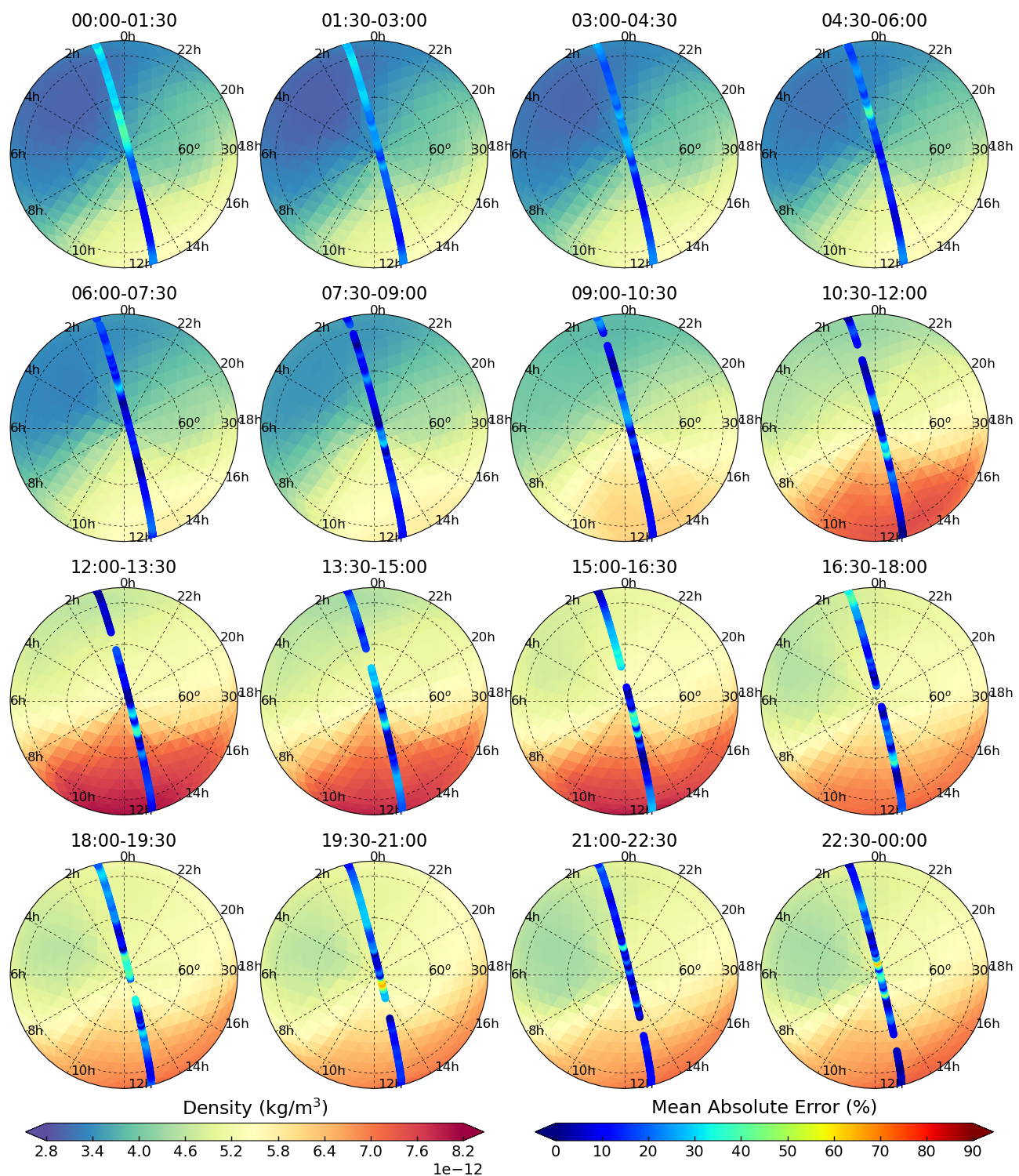}
    \caption{90 minute average density maps of the northern hemisphere on October 1, 2002. The line shows the flight path of CHAMP for that period with colors depicting the mean absolute error.}
	\label{f:stormN}
\end{figure}

\begin{figure}[htb!]
	\centering
	\small
	\includegraphics[width=1\textwidth]{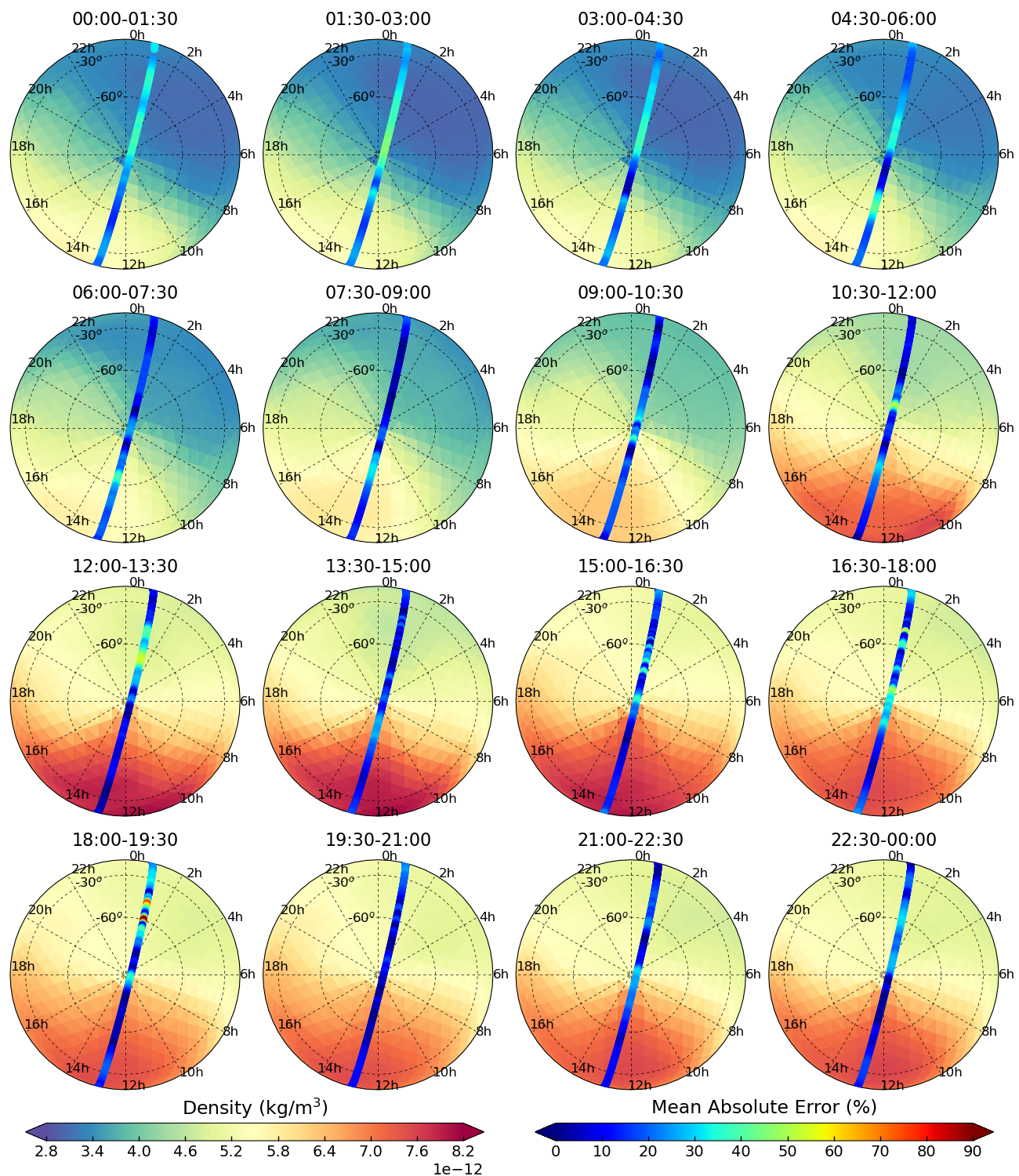}
    \caption{90 minute average density maps of the southern hemisphere on October 1, 2002. The line shows the flight path of CHAMP for that period with colors depicting the mean absolute error.}
	\label{f:stormS}
\end{figure}

We demonstrate global storm predictions by looking October 1, 2002 where \textit{Dst} reached a minimum value of -176\textit{nT}. The model inputs are averaged in 90 minute segments and an array of location inputs are provided for each of the 16 space weather conditions. The locations used are those used by the EXTEMPLAR model, but any locations can be input. The altitude was fixed at 412 km which was the average altitude of CHAMP during this period. We generated the global density maps (which represent the 90 minute average density distribution) and plot them with the mean absolute error along the CHAMP orbit. This is displayed in Figures \ref{f:stormN} and \ref{f:stormS} showing the northern and southern hemispheres, respectively. Note: the direct probability model was used here for ease of evaluation as this is simply a demonstration of global prediction.

The maps in both figures show that even though the training data is local, the model is able to provide reasonable global representations of thermospheric mass density. The diurnal structure of the thermosphere is highlighted in the the first four segments, prior to the onset of the storm. The \textit{F\textsubscript{10}} and \textit{F\textsubscript{81c}} values for this day are 139.7 and 173.9 sfu, respectively. This indicates the thermosphere is highly driven by the temperature variations which is evident in both figures. The southern hemisphere (Figure \ref{f:stormS}) displays a more significant impact of the storm as the high density values flow into the night-side at high latitudes. In the actual CHAMP data, the densities can often be erratic, particularly during storms, as a result of imperfect data processing schemes. However, these figures show that the model is able to learn the overall relationship of the inputs while being robust to outliers. This is emphasized by the continuity and smooth transitions between high- and low- density regions.

In terms of the model performance, the errors are lower in the northern hemisphere (16.84\%) than in the southern hemisphere (17.19\%). However, this could be a result of CHAMP's location during certain periods of the storm. The maximum errors (seen in dark red) are all in the southern hemisphere. There is also an interesting trend of higher errors in the night-side. This is because the density values are lowest here, so deviations from the true value have a stronger impact on the percent error. The error is shown as a percent for normalization as the density ranges by a factor of three as a function of location and time. It is important to reiterate that the density encountered by CHAMP at any point is not necessarily the density shown on the map as these are averages.

\end{document}